\documentclass{article}

\usepackage{placeins}
\usepackage{colortbl}
\usepackage{listings}
\usepackage{wrapfig}

\usepackage{amsmath,amssymb}
\usepackage{graphicx}
\usepackage{xcolor}
\usepackage{placeins}
\usepackage{colortbl}

\usepackage{microtype}
\usepackage{graphicx}

\usepackage{booktabs} 
\usepackage{hyperref}
\usepackage{wrapfig}
\usepackage{soul}
\usepackage[numbers]{natbib}
\usepackage{authblk}
\usepackage[T1]{fontenc}    
\usepackage{nicefrac}       
\usepackage{amsmath,amsthm}
\usepackage{amssymb}
\usepackage{mathtools}
\usepackage{algorithm}
\usepackage{color}
\usepackage{subcaption}
\usepackage{fullpage}
\usepackage{algpseudocode}
\usepackage{bbm}
\usepackage{caption}

\usepackage[capitalize,noabbrev]{cleveref}

\theoremstyle{plain}

\theoremstyle{definition}

\theoremstyle{remark}

\usepackage{color-edits}
\addauthor{st}{blue}
\addauthor{lpf}{green}
\newcommand{\inline}[1]{{\small $#1$}}
\usepackage[utf8]{inputenc} 
\usepackage{url}            
\usepackage{amsfonts}       
\usepackage{xcolor}         

\definecolor{codegreen}{rgb}{0,0.6,0}
\definecolor{codegray}{rgb}{0.5,0.5,0.5}
\definecolor{codepurple}{rgb}{0.58,0,0.82}
\definecolor{backcolour}{rgb}{0.95,0.95,0.92}

\lstdefinestyle{mystyle}{
    backgroundcolor=\color{backcolour},   
    commentstyle=\color{codegreen},
    keywordstyle=\color{magenta},
    numberstyle=\tiny\color{codegray},
    stringstyle=\color{codepurple},
    basicstyle=\ttfamily\footnotesize,
    breakatwhitespace=false,         
    breaklines=true,                 
    captionpos=b,                    
    keepspaces=true,                 
    numbers=left,                    
    numbersep=5pt,                  
    showspaces=false,                
    showstringspaces=false,
    showtabs=false,                  
    tabsize=2
}

\title{Merge and Conquer: Evolutionarily Optimizing AI for 2048}
\author{
    Maggie Bai\thanks{Correspondence to: Maggie Bai \texttt{<maggie.y.bai@gmail.com>}}, 
    Ava Kim Cohen, 
    Eleanor Koss, 
    Charlie Lichtenbaum\thanks{Authors are listed alphabetically; all authors contributed equally.}
}
\begin{document}
\maketitle

\begin{abstract}    
Optimizing artificial intelligence (AI) for dynamic environments remains a fundamental challenge in machine learning research. In this paper, we examine evolutionary training methods for optimizing AI to solve the game 2048, a 2D sliding puzzle. 2048, with its mix of strategic gameplay and stochastic elements, presents an ideal playground for studying decision-making, long-term planning, and dynamic adaptation. We implemented two distinct systems: a two-agent metaprompting system where a ``thinker" large language model (LLM) agent refines gameplay strategies for an ``executor" LLM agent, and a single-agent system based on refining a value function for a limited Monte Carlo Tree Search. We also experimented with rollback features to avoid performance degradation. Our results demonstrate the potential of evolutionary refinement techniques in improving AI performance in non-deterministic environments. The single-agent system achieved substantial improvements, with an average increase of 473.2 points per cycle, and with clear upward trends (correlation $\rho$=0.607) across training cycles. The LLM's understanding of the game grew as well, shown in its development of increasingly advanced strategies. Conversely, the two-agent system did not garner much improvement, highlighting the inherent limits of meta-prompting.
\end{abstract}


\section{Introduction}
2048, a simple 2D sliding game, has gained immense popularity due to its engaging gameplay and deceptively complex strategy despite its simple mechanics. The game involves sliding numbered tiles on a 4×4 grid, merging same-value tiles with the ultimate goal of creating a tile with the value of 2048. While players control the movement of the tiles, new tiles appear randomly each move with values of 2 (90\% of the time) or 4 (10\% of the time), meaning that players must constantly adapt. This unpredictability means that optimal play requires both probabilistic reasoning and adaptability. 

LLMs have demonstrated remarkable abilities across a variety of reasoning and comprehension tasks. However, their aptitude for solving complex, strategic environments (such as chess, Go, or poker) is limited when no task-specific finetuning or guidance is provided \cite{wang2025}. This limitation prompts important questions about the nature of "improvement" in AI systems.

Since reaching 2048 requires significant depth and complexity in strategy, the game is a rich environment for studying machine-learning challenges in optimization and decision-making algorithms. Games are also long, requiring hundreds of moves to reach 2048, forcing any solver to use a reliable, reproducible strategy instead of relying on luck. Over the years, researchers have applied traditional ML techniques such as expectimax, minimax, and deep reinforcement learning to explore optimal play in 2048 \cite{guei2022}\cite{nneonneo}.

In this paper, we investigate evolutionary training methods for optimizing AI to solve the game of 2048. We present two distinct systems: (1) a two-agent “metaprompting” framework, where a “thinker” LLM refines gameplay strategies for an “executor” LLM, and (2) a single-agent system that refines a value function for a limited Monte Carlo Tree Search. Crucially, our approach uses closed-source base models without fine-tuning or RL, thereby probing the extent to which purely prompt-based iterative improvement can enhance an LLM’s decision-making.

Our results show that the single-agent system achieved substantial gains, with an average increase of 473.2 points per training cycle and a strong positive correlation (\inline{\rho}=0.607) indicating consistent upward improvement. We further observe that the LLM’s strategic reasoning evolved over time, developing increasingly sophisticated gameplay tactics. Conversely, the two-agent system did not demonstrate significant gains, illustrating important limits of meta-prompting for self-refinement. By comparing these two evolutionary approaches, we shed light on the trade-offs and constraints of self-improvement methods when neither fine-tuning nor traditional RL is available.

\section{Related Works}

The intersection of artificial intelligence (AI) and game-play optimization has spurred advancements in decision-making algorithms \cite{silver2017} \cite{szubert2014}. Within them, there are contributions in AI optimization for the game of 2048, the application of Monte-Carlo Tree Search (MCTS), iterative training with rollback, and AI for games.

\subsection{Monte Carlo Tree Search}
Monte Carlo Tree Search (MCTS) has been extensively used in optimizing strategies for games that require sequential planning. This balance has positioned MCTS as a widely used technique in areas such as board games, video games, and decision making under uncertainty. Recent advancements have further enhanced MCTS by integrating deep learning techniques, enabling improved generalization and adaptation to complex scenarios. 

\subsection{Iterative Training}
Iterative training methods, particularly those combining optimization and regression prevention, can be used to directly improve LLM performance without human grading or input. One famous example of this is from 2017, when Silver et al. introduced AlphaGo Zero \cite{silver2017}, a reinforcement learning framework that achieved superhuman performance in the game of Go using iterative self-play and continuous optimization. Unlike its predecessor, AlphaGo, which relied on human expert games for initial training, AlphaGo Zero started with randomly initialized parameters and improved solely through self-play. Further, it uses a single neural network as well as a tree search to evaluate positions and moves without executing any Monte Carlo rollouts. This was done through a reinforcement learning program that resulted in significant improvements \cite{silver2017}.

Similarly, AlphaEvolve was recently presented in June 2025, where Novikov et al. {\cite{novikov2025}} used a combination of evolutionary computation and LLM-based coding. AlphaEvolve is an autonomous system that employs a system of LLMs to directly modify and improve on algorithms. Through an evolutionary process, it incorporates continuous feedback from one or more evaluators to progressively improve performance. This approach was described to have the potential to drive both scientific breakthroughs and practical advancements. To demonstrate its broad applicability, they applied AlphaEvolve to various critical computational challenges, and it outperformed current solutions for problems in math and computer science \cite{novikov2025}.

Another evolutionary training approach, the Darwin Gödel Machine (DGM), builds self-improving AI. DGMs use foundation models to propose it's own agentic code improvements. Experiments demonstrate that DGMs improve their performance as more computational resources are allocated \cite{zhang2025}.

The conceptual underpinnings of AlphaEvolve and DGM raise the question of what method of self-improvement is the most effective and reliable way for a model to improve itself while keeping the base model unchanged. Thus, in this paper, we explore iterative methodologies to evaluate  the comparative performance of two evolutionary frameworks in controlled environments emphasizing strategy, decision-making, and optimization under uncertainty. By examining the strengths and weaknesses of metaprompting and code-based refinement, we aim to identify which evolutionary strategy yields better outcomes in terms of adaptability and reliability.

\subsection{AI Optimization for 2048}
The game of 2048 is an effective benchmark for AI due to its combination of deterministic mechanics and random tile generation, presenting challenges in long-term planning. Traditional methods for solving 2048 often rely on heuristic-based techniques. As an example, Kohler et al. \cite{kohler2019} in 2019 tested the efficacy of basic heuristics in a 2048 solver. They designed a system that starts with a set of predefined heuristic evaluation functions, created new compositions of these functions based on guidelines, and then evaluated them by running a number of games. Ultimately, they found that value functions that maximize empty spaces and monotonicity of tiles perform the most optimally out of other compositions.

Further, the application of reinforcement learning (RL) to 2048 demonstrates promising results. Szubert and Jaśkowski (2014) created a game-playing agent based on N-tuple networks alongside reinforcement learning methods \cite{szubert2014}. The experiments showed that the learning algorithm using afterstate-value functions yields agents capable of winning over 97\% of games.


\section{Methodology}

\subsection{Baseline}
We began with the simplest possible system---an LLM playing with only its training data as knowledge---to establish a baseline understanding of performance before implementing improvements. We began with Claude 3.7 Sonnet, as we believed that its more extensive thinking capabilities would improve performance. However, this was not the case, as Claude consistently achieved only a 512 tile before losing. This approach was also flawed in the temporal sense; each game took well over an hour to complete, making repeated trials and experimentation difficult. As such, we switched to a non-thinking LLM, GPT-4o, and allocated a smaller token budget. While this did increase the speed of play to a few minutes per game, performance dropped as well, and this system achieved a maximum tile of 256.

\subsection{Metaprompting}

Metaprompting was selected as our self-improvement paradigm due to its simplicity, ease of implementation, and demonstrated reliability in producing iterative improvements \cite{hiraou2024}. Compared to more complex self-improvement systems, such as reinforcement learning via external reward signals, metaprompting allows for lightweight adaptation within constraints of resource availability and experimental execution time. Based on the results of the baseline play, we decided to split the system into two separate agents. One LLM, GPT-4o, would be the ``executor," tasked with completing the actual moves during gameplay based on a written strategy. The other LLM, Claude 3.7 Sonnet, would be the ``thinker," and design new strategies between games that would then be included as part of the prompt for GPT-4o. 

For reliability, we also switched to running 20 games in parallel with a singular strategy. After the completion of all of the games, the thinker was given the game histories and reasoning from the executor, which it used to refine the strategy. This process of gameplay to refinement was repeated for 25 rounds. 

\clearpage

\subsection{MCTS Refinement}

\begin{wrapfigure}{r}{0.3\textwidth}
    \centering
    \includegraphics[width=\linewidth]{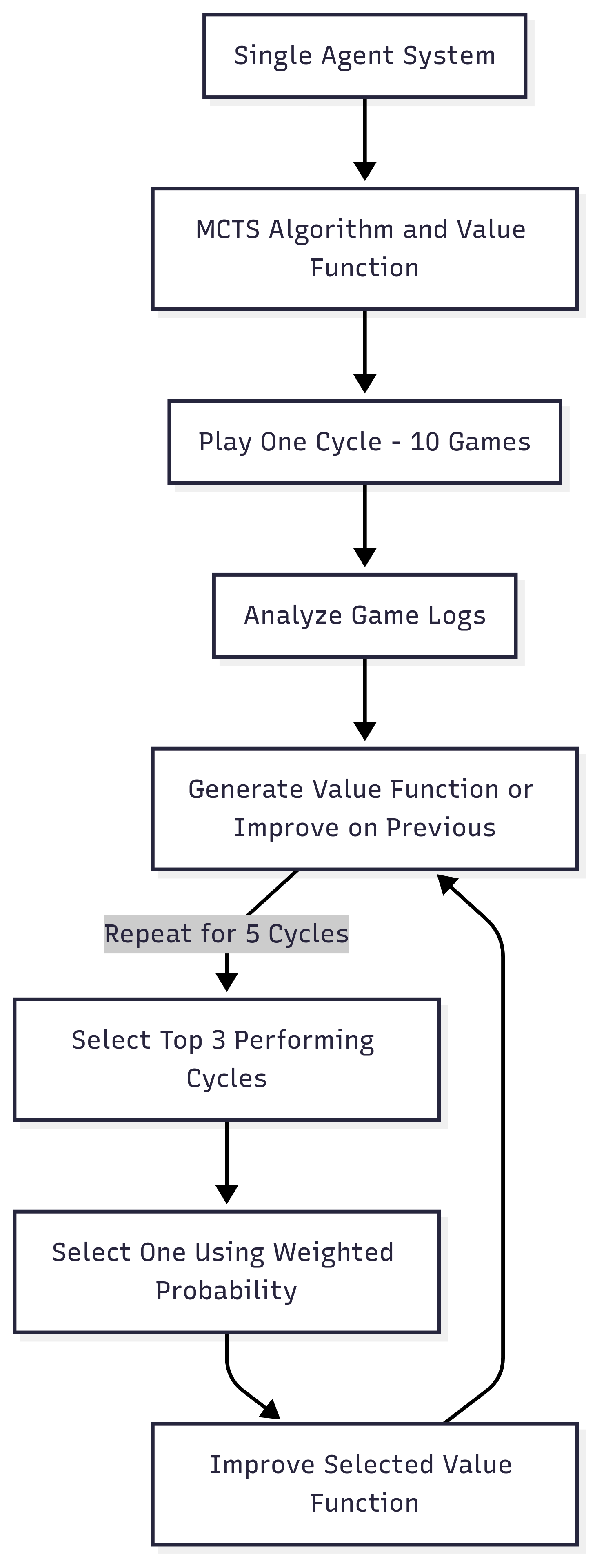}
    \caption{Program Flow of the MCTS Refinement System}
    \label{fig:system_architecture}
\end{wrapfigure}

We also decided to try a more programmatic approach, asking a singular model, Claude 3.7 Sonnet, to output code instead of verbal strategy. To accomplish this, we began with an MCTS playout system, as an unrefined system contains many avenues for refinement. Our chosen method of refinement was via an evaluation function for boards, as it would allow us to chart the LLM's understanding of the game itself, rather than straight coding. 

The initial setup was similar to the metaprompting one: we ran each value function produced by Claude for 10 games, with MCTS guiding the moves. After the set of 10 games, Claude analyzed the game histories and the latest generated function, and output a revised Python function for the next cycle. This function was directly stored in a .py file and executed, thereby granting the LLM a limited coding environment. 

After noticing in initial trials that the model would improve only to then deteriorate rapidly in the following cycles, we designed a rollback feature to maintain at least constant performance. To do this, we split the training into 5-cycle segments. Within those segments, the model would 

\noindent refine a value function, as specified above. After every five cycles, one of the previous value functions was selected using weighted probabilities based on average game score. The LLM was then given that function to improve on for the next five cycles, and the other four were discarded. The full program flow can be seen in Figure~\ref{fig:system_architecture}.



\section{Results}

\subsection{Two-Agent System}

\begin{figure}[h!]
     \centering
     \includegraphics[width=0.6\textwidth]{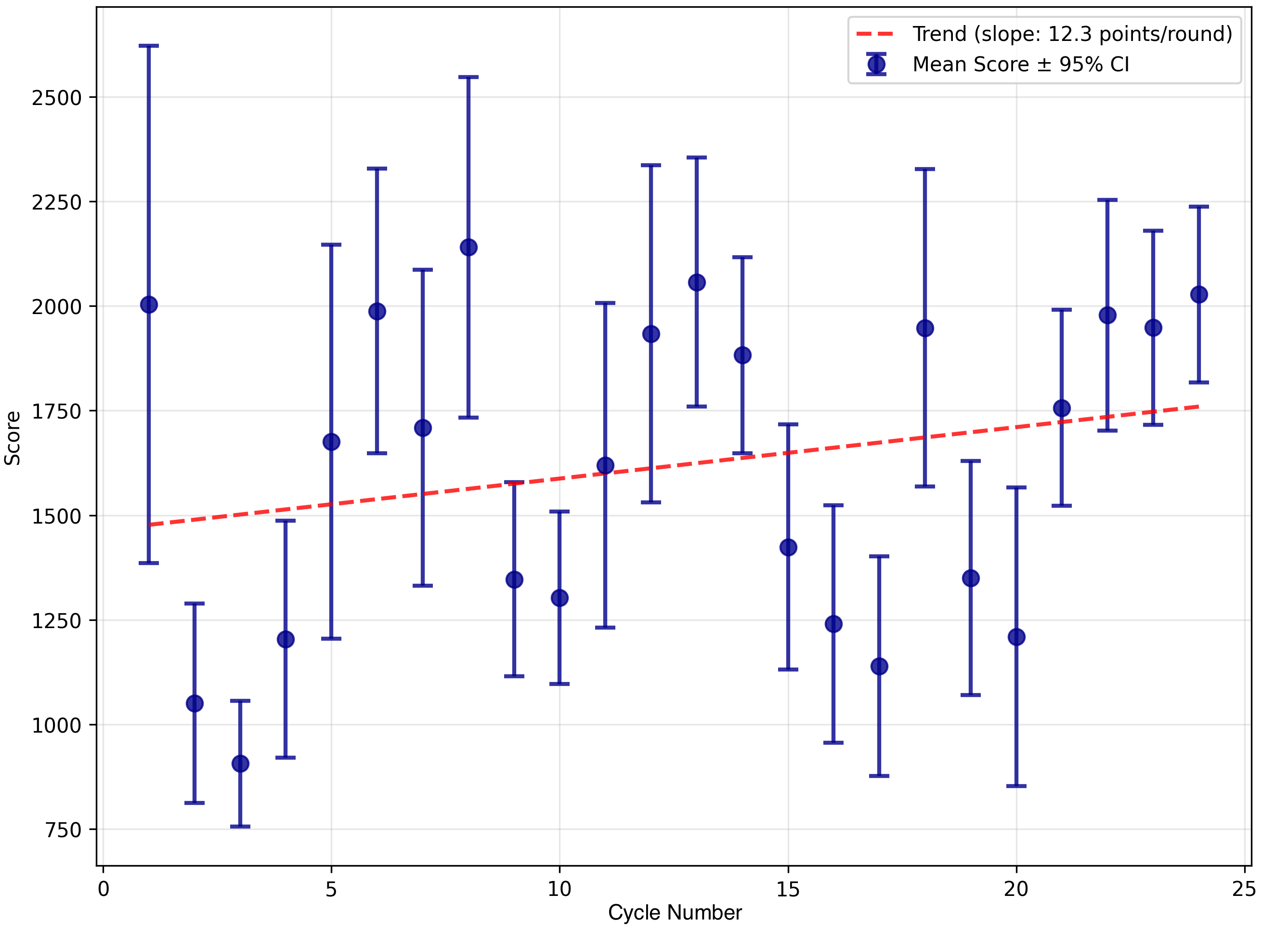}
     \caption{Metaprompting system, Average Score vs. Cycle Number; average score per round with 95\% confidence intervals over cycle progression. The red dashed line represents the trend (slope: 12.3 points/round)}
     \label{fig:two_agent_by_cycle}
\end{figure}

To evaluate whether the metaprompting system would yield improvements, we ran 25 cycles of evolving prompts, with 20 games in each round. The results are shown in Figure \ref{fig:two_agent_by_cycle}. This graph shows the relationship between round number and score. The blue dots represent the mean score in each round, with bars indicating the 95\% confidence intervals. The red dashed line represents the overall trend, which shows a small positive slope (12.3 points per round), but the wide confidence intervals suggest that individual round scores fluctuate significantly.


The results presented in both graphs suggest that the implemented strategy lacks effectiveness, as there is no consistent evidence of improvement or degeneration in rounds or games. The observed variability in performance, represented by large confidence intervals and the significant spread of individual game outcomes, implies that the results may be influenced more by random fluctuations than by any systematic application of the strategy. Furthermore, the lack of a significant upward or downward trend in performance suggests that this strategy does not produce tangible benefits in its current form.

\FloatBarrier

\subsection{One-Value Function}

\subsubsection{Changing the Value Function}

\begin{figure}[h!]
     \centering
     \includegraphics[width=0.7\textwidth]{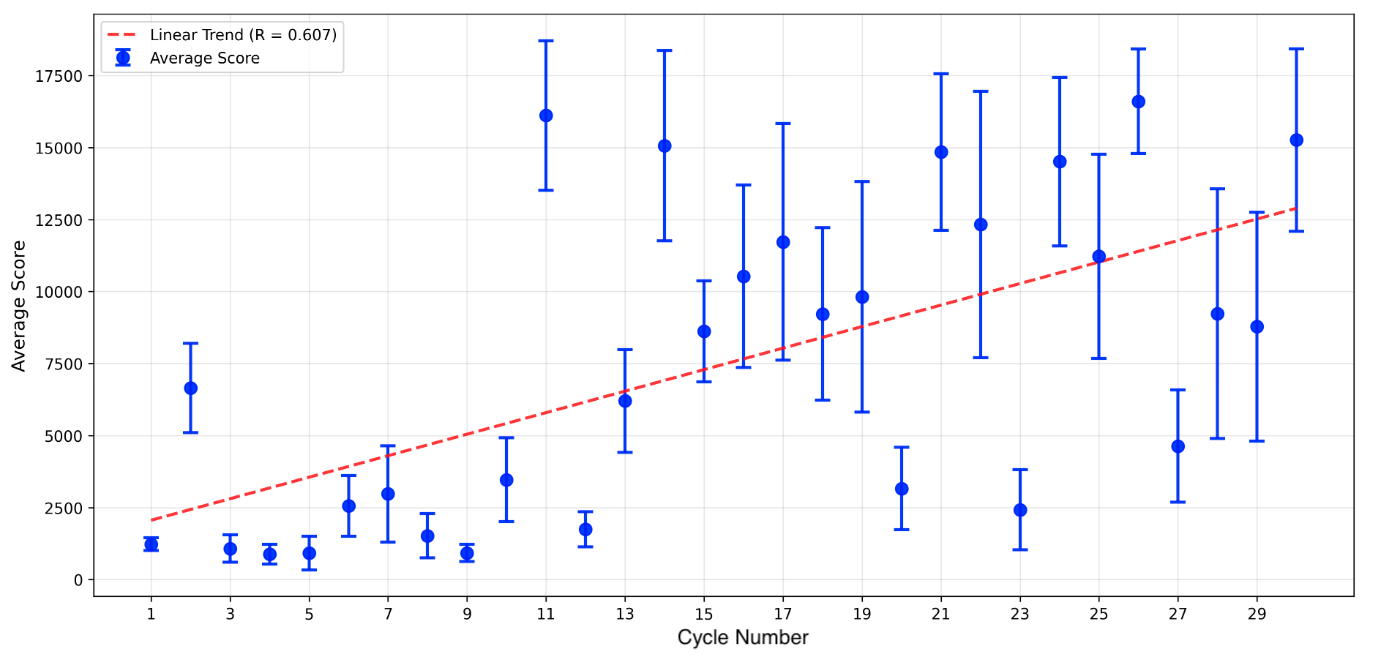}
     \caption{Altered Value Function with Rollback, Average Score vs. Cycle Number; average score across training cycles with 95\% confidence intervals. The red dashed line highlights a positive linear trend (\inline{\rho} = 0.607), indicating performance improvement as training progresses.}
     \label{fig:value_function_by_cycle}
\end{figure}

To test our MCTS and altering value function system, we ran 30 cycles of 10 games. Figure \ref{fig:value_function_by_cycle} summarizes the performance improvements observed during training of the 2048 game AI with MCTS reward tweaks and altered value functions that incorporate rollback mechanisms. 

\FloatBarrier

\textbf{Average Score vs. Cycle Number (Fig.~\ref{fig:value_function_by_cycle})}
Figure \ref{fig:value_function_by_cycle} aggregates scores by training cycle, showing the average performance and associated 95\% confidence intervals. The red trend line demonstrates steady improvement across cycles, with a slope of 473.2 points per cycle. The average scores have an \inline{\rho} value of 0.607, indicating a strong positive correlation between training cycle and score. Unlike expected training patterns where variance diminishes as strategies stabilize, the variability (represented by the error bars) grows throughout later cycles. This result suggests that while average performance increases, it remains inconsistent from game to game. Such behavior may be attributed to the exploratory nature of the MCTS tweaks, which introduce variability during the optimization process. 

\begin{figure}[!h]
\centering
    \begin{tabular}{|c|c|c|}
    \hline
    \textbf{Highest Tile} & \textbf{Two-Agent Distribution} & \textbf{Value-Function Distribution} \\
    \hline
    2048 & 0.0\%  & 5.3\% \\
    1024 & 0.0\%  & 28.0\% \\
    512  & 1.5\%  & 21.3\% \\
    256  & 25.2\% & 14.7\% \\
    128  & 49.4\% & 16.0\% \\
    64   & 21.3\% & 9.7\% \\
    32   & 2.7\%  & 3.7\% \\
    16   & 0.0\%  & 1.0\% \\
    8    & 0.0\%  & 0.3\% \\
    \hline
    \end{tabular}
    \caption{Distribution of highest tiles.}
    \label{fig:highest_tiles}
\end{figure}

\begin{figure} [!h]
\centering
    \begin{tabular}{|c|c|c|}
    \hline
    \textbf{Score Range} & \textbf{Two-Agent Distribution} & \textbf{Value-Function Distribution} \\
    \hline
    25K+     & 0.0\%  & 0.0\% \\
    20K--25K & 0.0\%  & 8.7\% \\
    15K--20K & 0.0\%  & 19.0\% \\
    10K--15K & 0.0\%  & 14.7\% \\
    5K--10K  & 0.8\%  & 28.3\% \\
    0--5K    & 99.2\% & 29.3\% \\
    \hline
    \end{tabular}
    \caption{Score range distribution.}
    \label{fig:score_range}
\end{figure}

Figure \ref{fig:highest_tiles} compares the distribution of the highest tiles achieved in the 2048 game by two different methods: the Two-Agent Distribution and the Value-Function Distribution. The Two-Agent Distribution appears to achieve lower scores overall, with the most frequently reached tiles being 128 (49.4\%) and 256 (25.2\%). Notably, this approach does not reach the higher tiles of 1024 or 2048 at all. In contrast, the Value-Function Distribution demonstrates significantly better performance, with higher tile scores being more common. It achieves tiles like 1024 (28.0\%) and 512 (21.3\%) at a notable frequency, and 5.3\% of games reach the 2048 tile.

Figure \ref{fig:score_range} compares the score range distributions for the Two-Agent and Value-Function approaches in the 2048 game. The Two-Agent Distribution overwhelmingly results in low scores, with 99.2\% of games scoring between 0–5K and just 0.8\% reaching the 5K–10K range. It fails to achieve any scores above 10K. In contrast, the Value-Function Distribution performs significantly better, with scores spread across higher ranges. While 29.3\% of games still fall in the 0–5K range, 28.3\% reach 5K–10K, and higher ranges (10K–15K, 15K–20K, and 20K–25K) account for 42.4\% of games combined.

\section{Discussion}
The results of the metaprompting training process as described in Figure \ref{fig:two_agent_by_cycle} indicate that a meta-prompting system is ineffective in a non-deterministic setting. The process of an executor agent following specific strategic instructions via prompts (supplied by a second agent or a meta-prompting process) does not fully leverage the broader adaptability needed to excel in such stochastic environments. This limitation arises because pre-defined strategies, while effective in deterministic contexts, fail to accommodate the inherent randomness and variability of outcomes. 

Further, prompts generated by the thinker agent have inherent limitations that ultimately fail to improve executor actions. The thinker agent may oversimplify key details to focus on the overall structure of the problem, thus leaving out critical information, especially regarding nuanced situations or corner-cases. Conversely, the thinker agent may also over-complicate the prompt, adding details that are irrelevant to game-play or lack effect. As a result, the executor agent is prompted with excess information, de-emphasizing crucial details with unnecessary ones.

The results of the value-function training process, namely in Figure \ref{fig:value_function_by_cycle}, demonstrate the LLM's capability to achieve high scores and win at 2048 given time. The cycle performance metrics show clear upward trends, indicating that the proposed training modifications enable the AI to learn and refine strategies. Thus, a code-based evolutionary training algorithm is most effective in the context of games like 2048.

One notable observation is the abrupt shift in performance from before cycle 10 to after. In Figure \ref{fig:value_function_by_cycle}, the game performance shoots up significantly during cycle 11. To understand why there is such a significant shift, we analyzed the value functions before and after cycle 10, and noticed changes in the following heuristics:

\begin{enumerate}
    \item \textbf{Corner positioning}
    
Before cycle 10, corner placement was rewarded with static bonuses based on the tile's position:
\begin{lstlisting}[language=Python]
if highest_pos in [(3, 0), (3, 3)]:  # Bottom corners (preferred)
    corner_bonus = 1.0
\end{lstlisting}

By contrast, on cycle 11, the proximity metric was improved to favor the bottom-right corner, while also considering other corners:
\begin{lstlisting}[language=Python]
br_distance = (3 - highest_row) + (3 - highest_col)
br_proximity = 1.0 - (br_distance / 6)
corner_proximity = 0.7 * br_proximity + 0.3 * other_corners_proximity
\end{lstlisting}

    \item \textbf{Monotonicity}

Before cycle 10, the value function checked for monotonicity along the bottom row:
\begin{lstlisting}[language=Python]
left_decreasing = all(row[i] >= row[i+1] for i in range(3))
right_decreasing = all(row[i] <= row[i+1] for i in range(3))
if left_decreasing or right_decreasing:
    monotonicity_score += 0.5
\end{lstlisting}

On cycle 11, however, snake pattern recognition was expanded to evaluate a zigzag structure across rows and columns:
\begin{lstlisting}[language=Python]
# Check bottom row (right to left decreasing)
for j in range(3):
    if board[3][3-j] > 0 and board[3][2-j] > 0:
        if board[3][3-j] >= board[3][2-j]:
            snake_score += 1
# Check inter-row connections
if board[3][0] > 0 and board[2][0] > 0 and board[3][0] >= board[2][0]:
    snake_score += 1
snake_ratio = snake_score / 15  # Normalize
\end{lstlisting}

    \item \textbf{Smoothness}

Smoothness, which evaluates adjacency of similar tiles for strategic merges, was entirely absent before cycle 10. After cycle 10, a smoothness factor was introduced:

\begin{lstlisting}[language=Python]
if j < 3 and board[i][j+1] > 0:
    diff = abs(math.log2(board[i][j]) - math.log2(board[i][j+1]))
    smoothness += 1 / (1 + diff)
smoothness_ratio = smoothness / 24  # Normalize
\end{lstlisting}

    \item \textbf{Weights}
    
The weights were also adjusted slightly before and after cycle 10.

Before:
\begin{lstlisting}[language=Python]
evaluation = (0.35 * empty_ratio +
             0.20 * highest_ratio +
             0.15 * corner_bonus +
             0.10 * bottom_row_ratio +
             0.10 * merge_value_ratio +
             0.05 * merge_ratio +
             0.05 * monotonicity_score) 
\end{lstlisting}

After:
\begin{lstlisting}[language=Python]
evaluation = (0.30 * empty_ratio +        # Reduced
             0.20 * highest_ratio +       # Reduced
             0.15 * corner_proximity +    # Improved metric
             0.10 * merge_ratio +         
             0.10 * smoothness_ratio +    # New metric
             0.15 * snake_ratio)          # Expanded metric
\end{lstlisting}

These adjustments reflect a small reduction in heuristic categories and a shift in focus toward structural and positional factors. By reducing the weight of simpler metrics such as empty tiles (from 35\% to 30\%) and highest tile ratio (from 25\% to 20\%), the new value function redistributed this weight to emphasize more nuanced evaluations, like corner proximity, smoothness, and snake patterns.

\end{enumerate}

Before cycle 10, the evaluation function prioritized simpler metrics, like empty tile ratio and basic corner bonuses, which supported early-game success but lacked advanced positional strategies. This led to consistent game development but difficulty in achieving higher tiles like 2048, as there was limited focus on maintaining ordered board structures. As such, we can see that more than just writing code, the model gained some understanding of the underlying strategy of the game itself. Additionally, this setup was not stuck in local optima, but managed to continue development into newer strategies. 

While our experiments showed a steady increase in performance, data also shows increasing variability. The variability observed in later training cycles highlights the need for techniques to stabilize learning outcomes. Future research could explore more advanced optimization options, such as alternate exploration-exploitation strategies to reduce inconsistencies.

\section{Conclusion}

This study explored optimizing artificial intelligence for 2048 using evolutionary training methods. Following the implementation of evolving value functions, our results showed steady increases in average scores with structured exploration and rollback strategies. On the contrary, our results show that a meta-prompting system with multiple agents has major limitations highly probabilistic settings. The ability to refine value functions intelligently through iterative processes positions these systems as effective tools for optimizing AI in similar games to 2048 and in similarly random contexts. Furthermore, the observed variability in later cycles suggests further avenues for exploration in stabilizing training outcomes and reducing inconsistency. 

\section*{Acknowledgment}
We would like to thank Jump Trading, as well as our mentors Lucas Baker, Nan Yang, Baihong Jin, and Loren Puchalla Fiore.

\bibliography{references}
\bibliographystyle{icml2024}
\end{document}